\documentclass[conference]{IEEEtran}

\usepackage{graphicx}
\usepackage{threeparttable}
\usepackage{epsfig}
\usepackage{epstopdf}
\usepackage[ruled,vlined,linesnumbered]{algorithm2e}
\usepackage{amssymb}
\usepackage{amsmath}
\usepackage{subfigure}
\usepackage[justification=centering]{caption}
\usepackage{amsfonts}

\usepackage{flushend}
\usepackage[hyphens]{url}
\usepackage{hyperref}
\hypersetup{breaklinks=true}

\usepackage{algpseudocode}

\ifCLASSINFOpdf

\else

\fi

\begin{document}
\title{Lightweight LSTM Model for Energy Theft Detection via Input Data Reduction}
\author{\IEEEauthorblockN{Caylum Collier}
\IEEEauthorblockA{School of Computer Science and Information Technology\\
University College Cork\\
Ireland\\
Email: caylumcollier77@gmail.com}
\and
\IEEEauthorblockN{Krishnendu Guha}
\IEEEauthorblockA{School of Computer Science and Information Technology\\
University College Cork\\
Ireland\\
Email: kguha@ucc.ie}
}

\markboth{Journal of \LaTeX\ Class Files,~Vol.~14, No.~8, August~2015}%
{Shell \MakeLowercase{\textit{et al.}}: Bare Demo of IEEEtran.cls for Computer Society Journals}

\IEEEtitleabstractindextext{
\begin{abstract}
With the increasing integration of smart meters in electrical grids worldwide, detecting energy theft has become a critical and ongoing challenge. Artificial intelligence (AI)-based models have demonstrated strong performance in identifying fraudulent consumption patterns; however, previous works exploring the use of machine learning solutions for this problem demand high computational and energy costs, limiting their practicality -- particularly in low-theft scenarios where continuous inference can result in unnecessary energy usage. This paper proposes a lightweight detection unit, or watchdog mechanism, designed to act as a pre-filter that determines when to activate a long short-term memory (LSTM) model. This mechanism reduces the volume of input fed to the LSTM model, limiting it to instances that are more likely to involve energy theft thereby preserving detection accuracy while substantially reducing energy consumption associated with continuous model execution. The proposed system was evaluated through simulations across six scenarios with varying theft severity and number of active thieves. Results indicate a power consumption reduction exceeding 64\%, with minimal loss in detection accuracy and consistently high recall. These findings support the feasibility of a more energy-efficient and scalable approach to energy theft detection in smart grids. In contrast to prior work that increases model complexity to achieve marginal accuracy gains, this study emphasizes practical deployment considerations such as inference efficiency and system scalability. The results highlight the potential for deploying sustainable, AI-assisted monitoring systems within modern smart grid infrastructures.
\end{abstract}

}

\maketitle
\IEEEdisplaynontitleabstractindextext
\IEEEpeerreviewmaketitle

\vspace{-10pt}
\section{Introduction}
\label{sec:introduction}
The smart grid is an increasingly integral component of modern energy infrastructure across the globe. One reason many countries are shifting over to smart grids, is the data provided by the infrastructure -- particularly smart meters -- which enables large-scale detection of irregularities in energy usage, including potential fraud\cite{HAQ2023634}. Consequently, a range of machine learning-based methods have been developed to identify energy theft based on this data. However, the volume of data generated by smart meters presents significant challenges. For instance, in Ireland, approximately 1.9 million smart meters \cite{esbSmartMeters} report data every half hour, which means each smart meter reports 48 times a day, therefore 1.9 million smart meters generate 91.2 million data points per day, or 638.4 million data points per week. Feeding this enormous quantity of data into a theft detection model demands substantial computational resources\cite{Zark2024}.

While various systems have been developed to detect energy theft, achieving detection accuracy rates as high as 92.5\% \cite{7434588}, they share a critical flaw: they typically process all data indiscriminately, regardless of the likelihood of fraudulent activity. Although this may be manageable in smaller-scale applications, the situation becomes economically infeasible in countries with large populations, such as India, with a population of 1.4 billion, the volume of smart meter data would render exhaustive processing impractical. The computational cost of processing this data could easily surpass the value of the stolen energy itself, rendering traditional detection solutions impractical.

Improving the system of theft detection to be more resource efficient would have the benefit of allowing for a wider adoption rate across the globe. The financial impact of energy theft can lead to increased cost of energy on consumers, and lower amounts of capital allocated to expanding and fortifying electrical infrastructure. Allowing for cheaper energy theft detection could help redistribute savings into critical areas, such as creating a more stable grid or lowering energy prices for consumers\cite{8281943}.

Existing energy theft detection solutions are often computationally expensive\cite{10429847}, which is worsened by the fact that they analyze consumption patterns for all users, despite the majority not engaging in theft\cite{su15064868}. This paper addresses this inefficiency by proposing a novel pre-filtering mechanism that activates an AI-based detection model only when suspicious activity is detected. This approach significantly reduces energy overhead, enhancing the feasibility of large-scale deployment in real-world smart grids. Additionally, the paper introduces a lightweight long short-term memory (LSTM) model with just two layers, offering a balance between pattern recognition capabilities and computational efficiency. The combined system -- a watchdog filter and a streamlined LSTM -- was evaluated through simulations across multiple theft scenarios to assess its impact on both energy savings and detection performance. The key contributions of this work are: (1) a watchdog mechanism that filters input to the detection model, and (2) a lightweight LSTM architecture tailored for efficient energy theft detection.

The remainder of this paper is structured as follows. Section~\ref{sec:background_and_motivation} discusses the structure of the grid, the mechanisms of energy theft, and reviews related work on energy theft detection. Section~\ref{sec:proposed_mechanism} describes the system architecture, including the watchdog mechanism and the theft detection model. Section~\ref{sec:methodology} outlines the implementation of the proposed system and the experimental setup. Section~\ref{sec:experimentation_and_results} presents the test scenarios and interprets the results. Finally, Section~\ref{sec:conclusion} concludes the paper.

\section{Background and Motivation}
\label{sec:background_and_motivation}
\subsection{Grid Topology}
\begin{figure}[ht]
  \centering
  \includegraphics[width=0.475\textwidth]{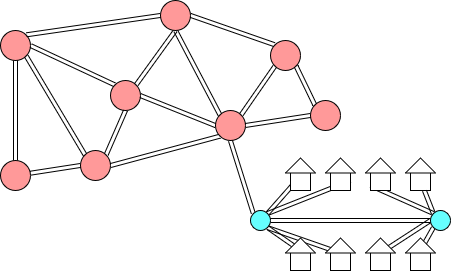}
  \caption[Low Level Grid Topology]{Simplified illustration of the grid structure.}
  \label{fig:low_grid}
\end{figure}

The electrical grid is structured into multiple layers based on voltage levels. At high and medium voltage levels (approximately 400 kV to 400 V), the grid exhibits a highly interconnected structure, where most nodes have multiple incoming and outgoing connections. In contrast, the lowest layer of the grid follows a radial topology, characterized by a single point of energy entry\cite{8810603}. This setup enables metering at the entry node, allowing precise measurement of the energy entering the radial section. This structure is illustrated in Figure~\ref{fig:low_grid}. Higher levels are highly interconnected, with nodes representing masts, substations, and similar infrastructure, while edges denote power lines. The lowest level, shown with smaller nodes, follows a radial design and connects directly to residential homes.

Such a structure is ideal for the proposed mechanism, as it facilitates the detection of discrepancies between the measured incoming energy and the aggregated consumption reported by individual smart meters within the radial segment.

\subsection{Energy Theft}
The goal of energy theft is to reduce the amount of money spent on energy. There are two primary methods by which energy theft is carried out: tampering with the smart meter to report lower usage than what was actually consumed, and bypassing the meter entirely by siphoning electricity from the grid before it is metered\cite{10429847}. Meter tampering may involve physical tampering or cyber-attacks aimed at altering the data it reports. In contrast, bypassing the meter usually requires illegal rewiring or tapping into nearby distribution lines. Both methods result in significant revenue losses for utilities and can lead to system imbalances, making theft detection a critical priority in smart grid operations. Additionally, such activities can pose safety risks due to unregulated and potentially hazardous modifications to the electrical infrastructure.

\subsection{Related Work}
\subsubsection{Machine Learning-Based Detection Approaches}
A wide range of machine learning models has been proposed for theft detection. Traditional classifiers, such as decision trees, support vector machines (SVM), and ensemble methods, have been widely used. Jindal et al. \cite{7434588} combined decision trees with SVMs to improve classification robustness. Nagi et al. \cite{4762604, 5738432} and Depuru et al. \cite{5772466, 6039050} also used SVMs and neural networks to detect irregular consumption behavior in developing regions.

More recent efforts apply deep learning to capture complex temporal patterns in consumption data. Chen et al. \cite{9061565} proposed a Deep Bidirectional Recurrent Neural Network (ETD-DBRNN), while Nabil et al. \cite{8875329} employed deep RNNs with multi-objective evolutionary tuning. Convolutional architectures have also been explored; for example, Zheng et al. \cite{8233155} introduced a wide and deep CNN to capture both global and local consumption features.

Hybrid pipelines further enhance performance albeit at the cost of greater complexity. Aldegheishem et al. \cite{9344652} combined SMOTE and WGAN for data balancing, AlexNet/GoogLeNet for feature extraction, and classifiers like LightGBM and AdaBoost. These approaches achieved high accuracy but incurred significant computational overhead -- highlighting the need for methods that can reduce input volume without compromising performance.

\subsubsection{Alternative and Specialized Techniques}
Some researchers have explored less conventional techniques. Jiang et al. \cite{1177814} applied wavelet-based feature extraction with ensemble classifiers. Ramos et al. \cite{5530391} introduced the Optimum-Path Forest (OPF) classifier, emphasizing robustness in noisy data. Nizar et al. \cite{4578740} utilized Extreme Learning Machines (ELM) and their online variants for faster pattern recognition.

Other studies focus on behavior-based or spatial anomaly detection. Jokar et al. \cite{7108042} used distribution transformer-level monitoring to identify suspicious regions with minimal sampling, while Cabral et al. \cite{1400905, 5275809} leveraged rough set theory and hybrid techniques like Self-Organizing Maps to detect theft under low-consumption scenarios.

\subsubsection{Relation to This Work}
The idea of using a “parent” or upstream meter to monitor regional consumption is conceptually aligned with the “central observer meter” proposed by Bandim et al. \cite{1335175}. However, their approach lacks machine learning components and performs poorly under low-resolution metering. In contrast, this work integrates the parent meter into a layered system that uses a lightweight watchdog mechanism to trigger AI-based detection only when necessary. This reduces computational and energy overhead, addressing a major limitation in prior solutions.

\subsubsection{Limitations of Previous Works}
While many prior studies report strong performance in terms of accuracy, precision, and recall, the underlying systems often involve complex architectures that incur significant computational and energy costs. These resource-intensive models may be impractical for continuous deployment, especially in large-scale settings. Introducing a lightweight solution -- such as a watchdog mechanism that activates the primary detection model only when suspicious activity is likely -- can reduce the operational costs associated with widespread energy theft detection.

\section{Proposed Mechanism}
\label{sec:proposed_mechanism}
\subsection{Architecture Design}
\begin{figure}[htbp]
  \centering
  \includegraphics[width=0.475\textwidth]{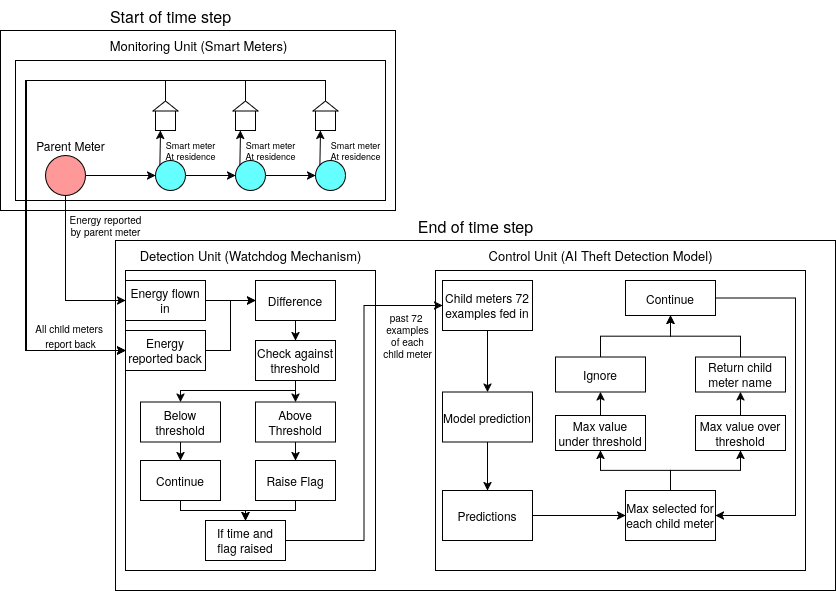}
  \caption[System Components]{Relationships between system components over a single timestep, segmented into duration (top) and end (bottom).}
  \label{fig:block_diagram}
\end{figure}

There are three primary components to the proposed mechanism: the monitoring unit -- which continuously records energy consumption; the detection unit -- which identifies areas of concern (also referred to as the watchdog); and the prediction unit -- which precisely identifies where energy theft occurs (the AI model). These components operate in a tightly coupled manner, with each module relying on the outputs of the others for full functionality. Figure~\ref{fig:block_diagram} illustrates this relationship.

The monitoring unit remains active throughout the entire timestep, continuously collecting data. The detection unit activates at the end of each timestep, once data from both parent and child meters has been aggregated. As the prediction unit depends on the detection unit’s output, it too operates at the end of the timestep.

\subsubsection{The Monitoring Unit}
The monitoring unit primarily leverages existing smart grid infrastructure -- specifically, residential smart meters that record and transmit energy consumption data. These child meters typically report usage to a centralized repository; in this system, they transmit data to an intermediary detection unit.

A minimal addition to the existing infrastructure is proposed in the form of parent meters -- functionally identical devices placed at choke points within the radial portion of the grid to record the total energy supplied to a group of child meters. This setup enables comparative analysis between the total input (from the parent meter) and the individual inputs (from the child meters) to identify potential theft.

\begin{figure}[htbp]
  \centering
  \includegraphics[width=0.475\textwidth]{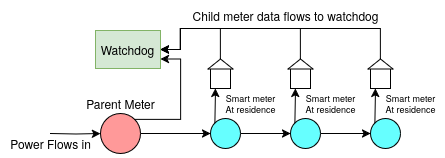}
  \caption[Structure of The Monitoring Unit]{The structure of the Monitoring Unit.}
  \label{fig:monitoring_unit_diagram}
\end{figure}

Energy flows through the parent meter on the way to the child meters. The consumption data from both the parent meters and child meters is then sent to the watchdog, where anomalies are flagged. The overall monitoring structure is illustrated in Figure~\ref{fig:monitoring_unit_diagram}.

\subsubsection{The Detection Unit}
\begin{figure}[htbp]
    \centering
    \includegraphics[height=0.475\textwidth]{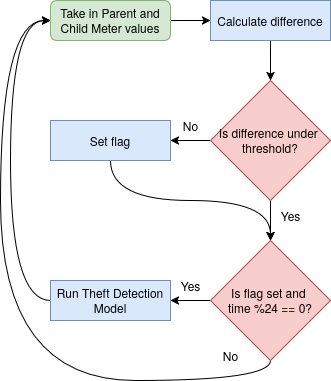}
    \caption[Flow Chart of The Detection Unit]{The flow chart for the detection unit.}
    \label{fig:flow_detection}
\end{figure}

The detection unit (watchdog mechanism) analyzes smart meter data to detect potential energy theft. Each parent meter is connected to a dedicated detection unit. At the end of each time interval, the detection unit receives input from the parent meter and all associated child meters. Additionally, it requires the estimated technical loss rate associated with the parent meter’s radial portion of the grid -- which is the rate at which energy is lost through friction in the form of heat, typically between 2-5\% -- and the threshold for allowable difference, expressed as a percentage. These values are provided directly to the detection unit and are not derived from the monitoring unit. The flow of the detection unit is depicted in Figure~\ref{fig:flow_detection}.

The algorithm in use for the detection unit calculates the difference between the energy flowing into the radial portion of the grid from a parent meter and the sum of reported energy consumption from its associated child meters, accounting for technical loss. It then checks whether this difference exceeds the given threshold; if so, it raises a flag. Finally, it checks whether the flag has been raised and whether the current time aligns with the designated evaluation time; if both conditions are met, it runs the prediction unit.

The technical loss is calculated using a formula specific to the radial portion of the grid, with values varying for each detection unit based on the grid’s characteristics. The threshold value determines the unit's sensitivity; a lower threshold results in higher sensitivity. Since the technical loss formula isn't always 100\% accurate, the threshold provides flexibility by allowing a maximum level of "allowable" loss.

The detection unit operates on a flexible reporting interval, which can be adjusted based on the reporting interval of the smart meters. At a designated time, such as 01:00 each night, the detection unit checks whether the flag has been raised. If the flag is set, the mechanism collects the previous 72 examples from each child meter connected to the parent meter and feeds them individually into the AI model. After processing the data, the flag is reset, and the watchdog mechanism begins the process anew, continuing this cycle until the next scheduled evaluation time.

The following variables are used in Algorithm~\ref{alg:detection_algorithm}.
\begin{itemize}
    \item $p[h]$: Energy (in kWh) reported by the parent meter at hour $h$.
    \item $c_i[h]$: Energy (in kWh) reported by child meter $i$ at hour $h$.
    \item $tl$: Technical loss, expressed as a decimal between 0 and 1.
    \item $d$: Computed discrepancy between parent and summed child meter readings.
    \item $n$: Number of child meters.
    \item $flag$: Indicator variable set when $d$ exceeds the threshold.
    \item $threshold$: Discrepancy percentage that triggers the flag.
    \item \texttt{control\_algorithm}: The algorithm used to classify potential theft.
    \item $inputs$: Input features passed to the prediction unit.
\end{itemize}

\begin{algorithm}[ht]
\caption{The Detection Unit Algorithm}
\label{alg:detection_algorithm}
\begin{tabbing}
\hspace{1.5em} \= \hspace{2em} \= \hspace{2em} \= \kill
\textbf{Loop forever:} \\
\> \textbf{for} $h$ in 0 to 23 \textbf{do} \\
\>\> $d \gets \left(\dfrac{p[h] - \left(\sum_{i=0}^{n} c_i[h] \cdot (1 + tl)\right)}{p[h]}\right) \cdot 100$ \\
\>\> \textbf{if} $d \geq \text{threshold}$ \textbf{then} \\
\>\>\> $flag \gets 1$ \\
\> \textbf{end for} \\
\> \textbf{if} $flag \geq 1$ \textbf{then} \\
\>\> $flag \gets 0$ \\
\>\> $outcomes \gets$ control\_algorithm(inputs) \\
\>\> \textbf{report} outcomes \\
\> \textbf{WaitForNextCycle()}
\end{tabbing}
\end{algorithm}

\subsubsection{The Prediction Unit}
The prediction unit is activated only when triggered by the detection unit. Without the detection unit, the system operates similarly to traditional theft detection approaches -- processing all smart meter data continuously and incurring substantial computational overhead. The prediction unit is modular, allowing the underlying model to be replaced should a more effective energy theft detection algorithm become available.

\begin{figure}[htbp]
    \centering
    \includegraphics[height=0.475\textwidth]{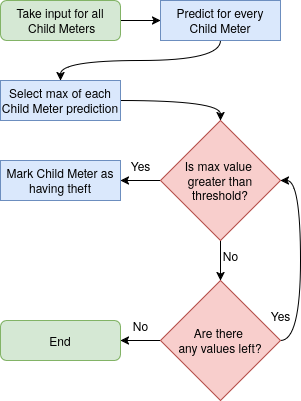}
    \caption[Flow Chart of The Prediction Unit]{This is the flow chart for the prediction unit.}
    \label{fig:flow_control}
\end{figure}

The prediction unit (AI theft detection model) detects energy theft on individual child meters. It receives the previous 72 data points for each child meter linked to a parent meter. For example, if the monitoring unit includes four child meters, the prediction unit processes four sets of 72 samples upon activation by the detection unit. These inputs are batch-processed to optimize computational efficiency. The model outputs a probability sequence of length 72 for each child meter, with values between 0 and 1. From each probability sequence, the maximum value is selected as the confidence score for theft. If this value exceeds the defined threshold (default 0.5), the system flags theft for that meter; otherwise, it reports no theft. The flow of the prediction unit is depicted in Figure~\ref{fig:flow_control}

The following variables are used in Algorithm~\ref{alg:prediction_algorithm}
\begin{itemize}
    \item $Child Meters$: A list of the previous $x$ examples for each child meter.
    \item $model$: The model in use to predict energy theft.
    \item $threshold$: The threshold above which theft is deemed to have occurred (default 0.5).
    \item $outcomes$: A list of either 1 or 0 representing whether theft has occurred(1) or not(0) for each child meter.
\end{itemize}

\begin{algorithm}[htbp]
\caption{The Prediction Unit Algorithm}
\label{alg:prediction_algorithm}
\begin{tabbing}
\hspace{1.5em} \= \hspace{2em} \= \hspace{2em} \= \kill
\textbf{Input:} Examples from child meters, trained model, threshold \\
\textbf{Output:} outcomes \\
\> \textbf{for each} child meter’s example set \textbf{do} \\
\>\> predictions $\gets$ model.predict(example) \\
\>\> \textbf{if} max(predictions) $\geq$ threshold \textbf{then} \\
\>\>\> outcomes.append(1) \\
\>\> \textbf{else} \\
\>\>\> outcomes.append(0) \\
\> \textbf{end for} \\
\textbf{return} outcomes
\end{tabbing}
\end{algorithm}

\subsection{LSTM-Based Detection Model}
An LSTM was chosen because of its ability to capture the sequential nature of the time-series data. LSTMs are a type of recurrent neural network (RNN) designed to handle long-term dependencies in sequences. They utilize gates that help retain relevant information over extended sequences and discard irrelevant data. This capability is essential for recognizing theft patterns in time series data, where past energy usage may influence future consumption behavior.

\begin{figure}[htbp]
  \centering
  \includegraphics[width=0.475\textwidth]{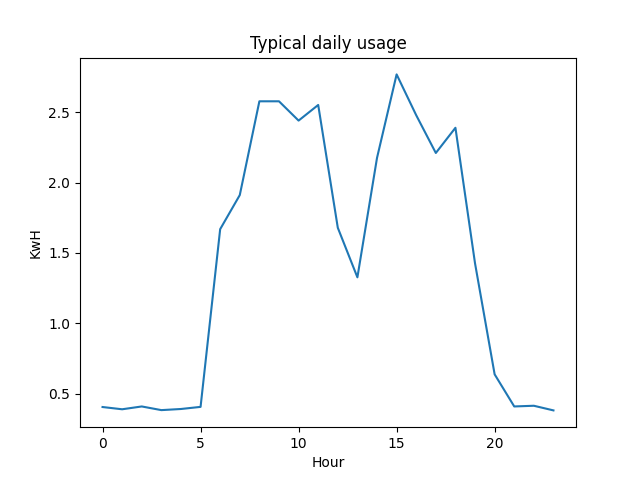}
  \caption[Typical Daily Energy Usage of a Residence]{This is an example of typical daily usage in a residence.}
  \label{fig:daily_kwh_usage}
\end{figure}

An example of typical daily energy usage for a residence is shown in Figure~\ref{fig:daily_kwh_usage}. This figure illustrates the consumption patterns that the long short-term memory (LSTM) model aims to learn and distinguish, particularly in the context of identifying anomalies such as energy theft. The data for this graph was sourced from a random example within the dataset and visualized using the PyPlot library.

The model consists of two main layers: a 64-neuron long short-term memory (LSTM) layer and a single-neuron dense output layer with a sigmoid activation function. This architecture is particularly lightweight for an LSTM, as it minimizes the number of parameters while maintaining the ability to effectively capture sequential patterns in the data. The choice of only a 64-neuron LSTM layer significantly reduces the computational load compared to deeper or more complex models that typically use hundreds or thousands of neurons. This makes the model highly efficient in terms of memory usage and processing power, which is crucial for large-scale deployment in real-world smart grids. Furthermore, the use of a simple dense output layer with a single neuron ensures that the model is not only computationally efficient but also minimizes the risk of overfitting, particularly when the dataset consists of relatively simple energy consumption patterns. The model outputs a probability at each timestep, representing the likelihood of energy theft occurring at that moment. The input data is typically a series of 72 time intervals, corresponding to 72 hours (or three days) of energy consumption data.

For each example, the LSTM layer outputs a sequence of probabilities, one for each of the input time steps. The model then selects the maximum probability from the input values, which corresponds to the time step in which the likelihood of energy theft is highest. If this maximum probability exceeds a predefined threshold (set to 0.5 in this model), the model classifies the instance as indicating potential energy theft for the corresponding child meter.

The primary input feature to the LSTM model is the energy consumption (kWh) at each time interval. The dropout layer with a rate of 0.5 was introduced to prevent overfitting. Given the relatively small amount of data, this regularization technique helps ensure that the model generalizes well without memorizing the training data. During model training, the data from all time steps of each child meter is fed as a batch to optimize resource utilization and speed up the training process.

\section{Methodology}
\label{sec:methodology}
\subsection{Dataset and Preprocessing}
A key challenge in electricity theft detection is the absence of standardized, publicly available datasets. This lack of benchmarks hinders meaningful comparison between studies. Real-world datasets also rarely include labeled theft instances, limiting the feasibility of supervised learning. The dataset used in this study, published by Quesada et al. \cite{quesada2024smartmeter} on Zenodo, contains per-meter CSV files with timestamped kWh consumption, imputation flags, and a metadata file with approximate location data. However, it lacks explicit theft examples. To address this, synthetic theft patterns were generated using six scenarios described by Zidi et al. \cite{ZIDI202313}, enabling evaluation under controlled conditions.

One of the key responsibilities of this section is the allocation of smart meters to parent meters. This is done via the \texttt{metadata.csv} file, which lists all meters and their corresponding data files. Meters are grouped based on a predefined maximum capacity per parent meter (200 in this case) and their ZIP code. The hypothesis here is that meters within the same ZIP code are more likely to exhibit similar energy usage patterns, which could make energy theft anomalies more detectable. This is also done to more closely simulate a real world situation, since the smart meters together in a radial portion of a grid would be in close proximity to each other.

\subsection{Synthetic Theft Generation}
Since the dataset lacks theft data, it must be synthetically generated. Due to the challenges in obtaining real-world energy theft data, the methods proposed by Zidi et al. \cite{ZIDI202313} are employed to simulate synthetic theft based on actual energy consumption patterns. Zidi et al. identify six distinct algorithms to model various types of energy theft, such as setting some values to zero or reducing the reported consumption by a certain fraction. These algorithms are adapted for the proposed system with a slight modification: the introduction of an additional parameter that controls the "severity" of the theft. This parameter, which is a value between 0 and 1, determines the likelihood that a specific theft algorithm will be applied to each data point. For example, if the severity parameter is set to 0.6, the function generates a random number between 0 and 1. If the generated number is less than 0.6, the corresponding theft algorithm is applied to the energy consumption value for that time period. This process is repeated for every time step in the dataset, and as a result, approximately 60\% of all generated examples will exhibit the specific type of theft corresponding to the severity value of 0.6.

\subsection{Simulation}
The simulation phase implements the detection unit as a function that follows the defined algorithm and internally encapsulates the prediction unit. This stage collects relevant outputs, including accuracy metrics and power consumption data, and writes them to corresponding result files. 

\begin{figure}[b]
\centering
\includegraphics[width=0.475\textwidth]{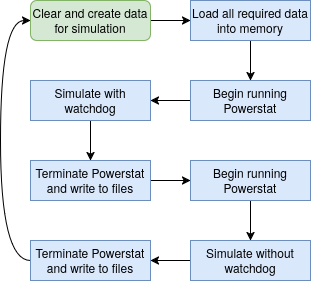}
\caption[Flow Chart of The Simulation]{This figure shows how the simulations were run in the form of a flow chart.}
\label{fig:simulation_flow}
\end{figure}

Two types of simulations were devised -- one with the detection unit and one without -- to evaluate whether the detection unit successfully reduces the power consumption of the AI-based theft detection model while only allowing a negligible amount of theft to go undetected. To measure power consumption, a wrapper function opens a background process and runs the Linux shell utility \texttt{powerstat}, immediately initiates the simulation, and then terminates the monitoring process. Simulation scenarios vary by theft rate and individual theft severity, configured using functions from the data pre-processing section. Each simulation uses 20\% of the dataset, representing one year’s worth of hourly data, and must process approximately 450 different parent meter and detection unit pairings. In a real-world setting, each parent meter would be paired with a single detection unit, but the scale used here enables measurement of global system behavior and resource consumption. The flow of the simulation process is depicted in Figure~\ref{fig:simulation_flow}.

Each scenario involves an element of randomness -- both in the selection of meter files and the specific time intervals during which theft occurs. However, both simulation modes (with and without the detection unit) are run on the same scenario to ensure fairness. This ensures that the stochastic nature of scenario generation does not bias the results toward either configuration.

The simulation returns two types of data: accuracy and power consumption. Power consumption is measured using \texttt{powerstat}, which records power usage in watts based on battery discharge rate throughout the execution. Accuracy results are divided into two types: watchdog accuracy (for the detection unit) and AI accuracy. The watchdog accuracy measures whether the watchdog activated within a 24-hour interval and whether theft occurred in that interval, to evaluate detection sensitivity.

Six distinct simulations were performed, varying two key dimensions: theft severity and number of thieves. Theft severity refers to how frequently an individual thief tampers with their energy consumption. In high-severity scenarios, a thief alters usage in 10\% to 95\% of time intervals, while in low-severity scenarios, the range is just 10\% to 20\%. The number of thieves represents the proportion of users who exhibit theft behaviors in each simulation. It is categorized as high (40\%), medium (20\%), or low (5\%). A summary of all combinations is shown in Table~\ref{tab:theft_matrix}.

\begin{table}[htbp]
\centering
\caption{Theft Simulation Scenarios Matrix}
\label{tab:theft_matrix}
\begin{tabular}{|c|c|c|c|}
\hline
\multicolumn{1}{|c|}{} & \multicolumn{3}{c|}{\textbf{Rate of Thieves}} \\
\cline{2-4}
\textbf{Theft Severity} & \textbf{40\%} & \textbf{20\%} & \textbf{5\%} \\
\hline
\textbf{0.10 -- 0.95\%} & sim\_hh & sim\_hm & sim\_hl \\
\hline
\textbf{0.10 -- 0.20\%} & sim\_lh & sim\_lm & sim\_ll \\
\hline
\end{tabular}
\end{table}

\section{Experimentation and Results}
\label{sec:experimentation_and_results}
\subsection{Power Consumption}

Table~\ref{tab:power_table} shows power usage across six simulations, with and without the watchdog. Simulation codes use two letters: the first indicates theft severity (high or low), and the second indicates thief count (many or few). For instance, \textit{sim hl} represents high theft severity with few thieves.

\begin{figure}[htbp]
\centering
\includegraphics[width=0.475\textwidth]{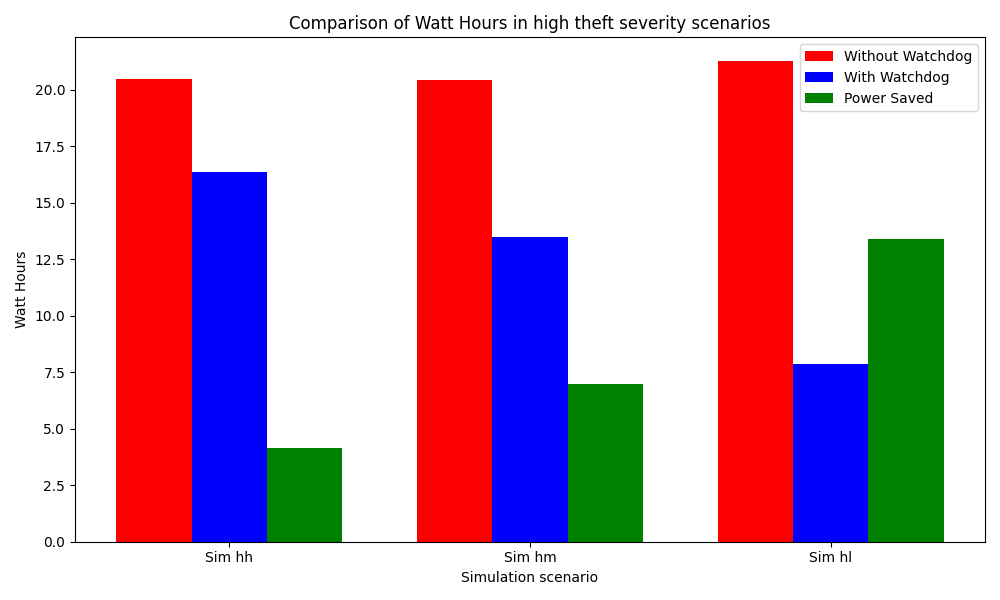}
\caption{Wh consumed with/without watchdog in high theft scenarios.}
\label{fig:results_power_high_sev}
\end{figure}

\begin{figure}[htbp]
\centering
\includegraphics[width=0.475\textwidth]{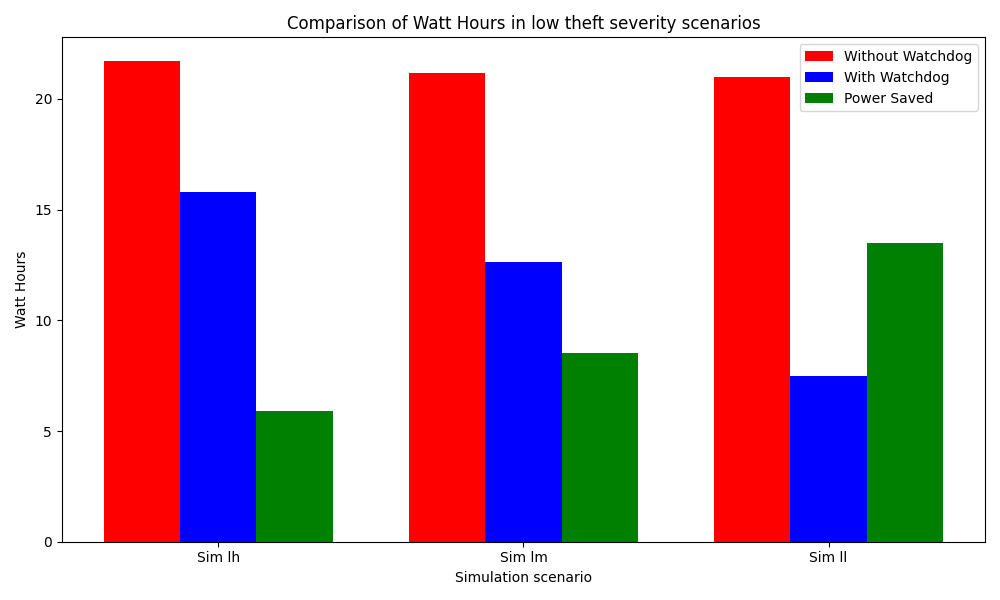}
\caption{Wh consumed with/without watchdog in low theft scenarios.}
\label{fig:results_power_low_sev}
\end{figure}

Figure~\ref{fig:results_power_high_sev} and Figure~\ref{fig:results_power_low_sev} plot Watt-hour consumption for high and low severity cases, respectively.

\begin{table}[htbp]
\caption{Power consumption with/without watchdog.}
\centering
\small
\setlength{\tabcolsep}{4.5pt}
\begin{tabular}{|c|c|c|c|c|c|c|c|}
\hline
\textbf{Sim} & \multicolumn{3}{c|}{\textbf{No Watchdog}} & \multicolumn{3}{c|}{\textbf{With Watchdog}} & \textbf{Diff} \\
\hline
 & \textbf{W} & \textbf{S} & \textbf{Wh} & \textbf{W} & \textbf{S} & \textbf{Wh} & \\
\hline
hh & 73799.5 & 4596 & 20.50 & 58912.8 & 3594 & 16.36 & 4.14 \\
lh & 78114.7 & 4823 & 21.70 & 56813.0 & 3489 & 15.78 & 5.92 \\
hm & 73586.8 & 4506 & 20.44 & 48524.3 & 2955 & 13.48 & 6.96 \\
lm & 76205.6 & 4688 & 21.17 & 45557.9 & 2779 & 12.65 & 8.52 \\
hl & 76546.6 & 4625 & 21.26 & 28281.6 & 1675 & 7.86 & 13.40 \\
ll & 75538.5 & 4645 & 20.98 & 26995.7 & 1615 & 7.50 & 13.48 \\
\hline
\end{tabular}
\label{tab:power_table}
\end{table}

Without the watchdog, simulations consistently use ~21 Wh. With the watchdog, power drops -- more significantly when more thieves are present -- due to reduced computation by filtering out events before AI processing.

\subsection{Model Accuracy}

\begin{figure}[htbp]
\centering
\includegraphics[width=0.475\textwidth]{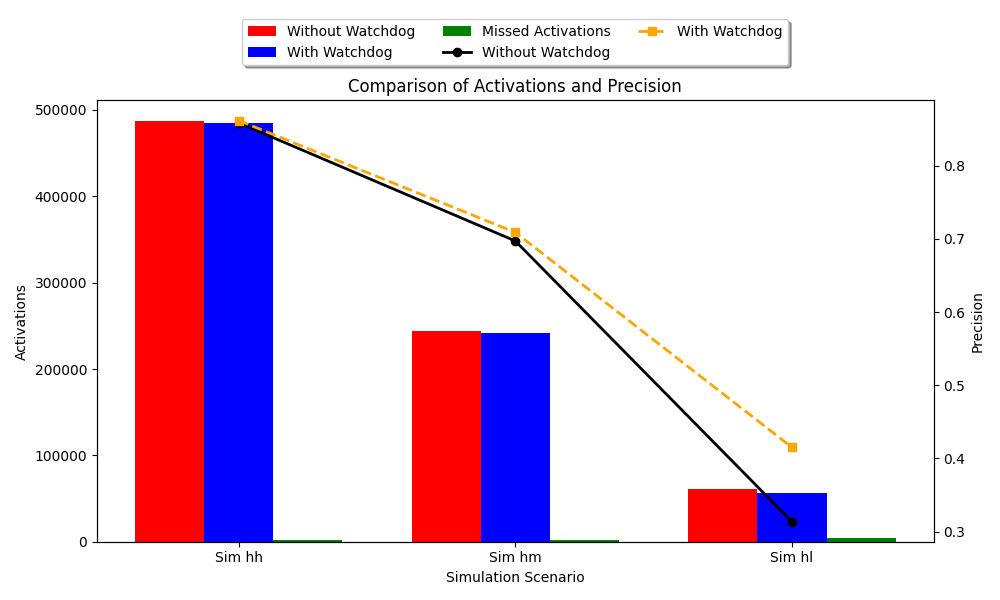}
\caption{Accuracy metrics for high theft scenarios.}
\label{fig:results_accuracy_high_sev}
\end{figure}

\begin{figure}[htbp]
\centering
\includegraphics[width=0.475\textwidth]{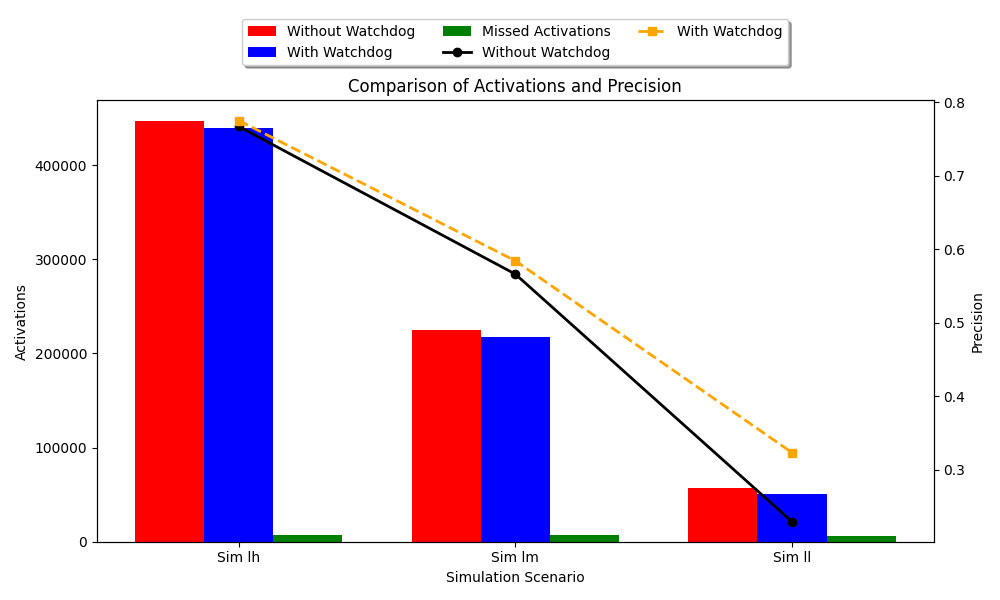}
\caption{Accuracy metrics for low theft scenarios.}
\label{fig:results_accuracy_low_sev}
\end{figure}

Table~\ref{tab:accuracy_table} shows precision, recall, and positive detection counts for all simulations. The final column shows the percentage of detections filtered out by the watchdog. Figure~\ref{fig:results_accuracy_high_sev} and Figure~\ref{fig:results_accuracy_low_sev} showcase the number of activations missed with/without the watchdog present, and the effect it has on precision for high and low theft scenarios respectively.

\begin{table}[htbp]
\caption{Model performance with/without watchdog.}
\centering
\small
\setlength{\tabcolsep}{4.5pt}
\begin{tabular}{|c|c|c|c|c|c|c|c|}
\hline
\textbf{Sim} & \multicolumn{3}{c|}{\textbf{No Watchdog}} & \multicolumn{3}{c|}{\textbf{With Watchdog}} & \textbf{Diff} \\
\hline
 & \textbf{P} & \textbf{R} & \textbf{Pos} & \textbf{P} & \textbf{R} & \textbf{Pos} & \\
\hline
hh & 0.8587 & 0.5842 & 486699 & 0.8620 & 0.5856 & 484700 & 0.41\% \\
lh & 0.7673 & 0.3953 & 446230 & 0.7754 & 0.3982 & 439408 & 1.53\% \\
hm & 0.6973 & 0.6055 & 244584 & 0.7090 & 0.6101 & 242018 & 1.05\% \\
lm & 0.5664 & 0.3940 & 225063 & 0.5845 & 0.3984 & 217525 & 3.35\% \\
hl & 0.3135 & 0.5788 & 60572 & 0.4151 & 0.6031 & 56461 & 6.79\% \\
ll & 0.2295 & 0.3966 & 57191 & 0.3235 & 0.4182 & 50805 & 11.17\% \\
\hline
\end{tabular}
\label{tab:accuracy_table}
\end{table}

\subsection{Results Analysis}
The results demonstrate that integrating the watchdog significantly reduces power consumption across all scenarios. This reduction is most pronounced when the number of thieves is high, which suggests that the watchdog effectively filters out unnecessary computations by reducing the number of inputs to the AI model. Notably, even in simulations with fewer thieves or lower theft severity, the watchdog consistently provides measurable energy savings.

In terms of accuracy, the impact of the watchdog on detection performance is minimal. Precision and recall values remain stable across simulations, indicating that the filtering process does not significantly degrade model effectiveness. However, the number of positive detections is slightly reduced, especially in scenarios with fewer total theft events. This trade-off is acceptable given the substantial power savings achieved.

Overall, the watchdog serves as an efficient pre-processing layer, striking a balance between computational cost and detection accuracy. It is particularly beneficial in environments with high-frequency theft, where the energy saved is maximized while maintaining reliable detection.

\section{Conclusion}
\label{sec:conclusion}
The primary contribution of the proposed system is a novel solution to improve the energy efficiency of existing AI-based energy theft detection models. By introducing a detection unit that performs lightweight calculations to filter potential signs of theft, the system significantly reduces the computational load on the AI model. This approach allows for lower power consumption without a major sacrifice in detection accuracy.

In addition to its power-saving benefits, the system offers flexibility. Various parameters, such as the threshold for activating the AI model, can be adjusted to optimize performance for different situations. Moreover, the design allows for easy replacement of the AI theft detection model, enabling the integration of more accurate or efficient models as they become available.

Ultimately, the proposed system offers a practical and scalable approach to reducing energy consumption in theft detection systems, providing significant benefits to real-world applications like smart grids, utility management, and energy monitoring.

\bibliographystyle{IEEEtran}
\bibliography{references}

\end{document}